\documentclass[runningheads,a4paper]{llncs}
\usepackage{bbding}
\usepackage{amssymb}
\usepackage{graphicx}
\usepackage{amsmath}
\usepackage{longtable}
\usepackage{float}
\usepackage{subfigure}

\usepackage{url}
\urldef{\mailsa}\path|mohanlin@ict.ac.cn|
\newcommand{\keywords}[1]{\par\addvspace\baselineskip
\noindent\keywordname\enspace\ignorespaces#1}

\begin{document}
\mainmatter
\title{Shape-Color Differential Moment Invariants under Affine Transformations}
\titlerunning{Shape-Color Differential Moment Invariants under Affine Transform}
\author{Hanlin Mo$^{1}$$^{(}$\Envelope$^{)}$$^{,}$%
\thanks{Student is the first author.}%
\and Shirui Li$^{1}$\and You Hao$^{1}$\and Hua Li$^{1}$}
\authorrunning{H.L. Mo et al.}
\institute{$^1$Key Laboratory of Intelligent Information Processing,\\Institute of Computing Technology, Chinese Academy of Sciences, Beijing, China\\
\mailsa\\}
\toctitle{Lecture Notes in Computer Science}
\tocauthor{Authors' Instructions}
\maketitle
\begin{abstract}
We propose the general construction formula of shape-color primitives by using partial differentials of each color channel in this paper. By using all kinds of shape-color primitives, shape-color differential moment invariants can be constructed very easily, which are invariant to the shape affine and color affine transforms. 50 instances of SCDMIs are obtained finally. In experiments, several commonly used color descriptors and SCDMIs are used in image classification and retrieval of color images, respectively. By comparing the experimental results, we find that SCDMIs get better results.
\end{abstract}

\section{Introduction}
Image classification and retrieval for color images are two hotspots in pattern recognition. How to extract effective features, which are robust to color variations caused by the changes in the outdoor environment and geometric deformations caused by viewpoint changes, is the key issue. The classical approach is to construct invariant features for color images. Moment invariants are widely used invariant features.

Moment invariants were first proposed by Hu\cite{1} in 1962. He defined geometric moments and constructed 7 geometric moment invariants which were invariant under the similarity transform(rotation, scale and translation). Researchers applied Hu moments to many fields of pattern recognition and achieved good results\cite{2,3}. Nearly 30 years later, Flusser et al.\cite{4} constructed the affine moment invariants (AMIs) which are invariant under the affine transform. The geometric deformations of an object, which are caused by the viewpoint changes, can be represented by the projective transforms. However, general projective transforms are complex nonlinear transformations. So, it's difficult to construct projective moment invariants. When the distance between the camera and the object is much larger than the size of the object itself, the geometric deformations can be approximated by the affine transform. AMIs have been used in many practical applications, such as character recognition\cite{5} and expression recognition\cite{6}. In order to obtain more AMIs, researchers designed all kinds of methods. Suk et al.\cite{7} proposed graph method which can be used to construct AMIs of arbitrary orders and degrees. Xu et al.\cite{8} proposed the concept of geometric primitives, including distance, area and volume. AMIs can be constructed by using various geometric primitives. This method made the construction of moment invariants have intuitive geometric meaning.

The above-mentioned moment invariants are all designed for gray images. With the popularity of color images, the moment invariants for color images began to appear gradually. Researchers wanted to construct moment invariants which are not only invariant under the geometric deformations but also invariant under the changes of color space. Geusebroek et al.[9] proved that the affine transform model was the best linear model to simulate changes in color resulting from changes in the outdoor environment. Mindru et al.\cite{10} proposed moment invariants which were invariant under the shape affine transform and the color diagonal-offset transform. The invariants were constructed by using the related concepts of Lie group. Some complex partial differential equations had to be solved. Thus, the number of them was limited and difficult to be generalized. Also, Suk et al.\cite{11} put forward affine moment invariants for color images by combining all color channels. But this approach was not intuitive and did not work well for the color affine transform. To solve these problems, Gong et al.\cite{12,13,14} constructed the color primitive by using the concept of geometric primitive proposed in \cite{8}. Combining the color primitive with some shape primitives, moment invariants that are invariant under the shape affine and color affine transforms can be constructed easily, which were named shape-color affine moment invariants(SCAMIs). In \cite{14}, they obtained 25 SCAMIs which satisfied the independency of the functions. However, we find that a large number of SCAMIs with simple structures and good properties are missed in \cite{14}.

In this paper, we propose the general construction formula of shape-color primitives by using partial differentials of each color channel. Then, we use two kinds of shape-color primitives to construct shape-color differential moment invariants(SCDMIs), which are invariant under the shape affine and color affine transforms. We find that the construction formula of SCAMIs proposed in \cite{14} is a special case of our method. Finally, commonly used image descriptors and SCDMIs are used for image classification and retrieval of color images, respectively. By comparing the experimental results, we find that SCDMIs proposed in this paper get better results.

\section{Related Work}
In order to construct image features which are robust to color variations and geometric deformations, researchers have made various attempts. Among them, SCAMIs proposed in \cite{14} are worthy of special attention. SCAMIs are invariant under the shape affine and color affine transforms. Two kinds of affine transforms are defined by
\begin{equation}
\left(
\begin{array}{c}
x^{'}\\
y^{'}\\
\end{array}
\right)
=SA \cdot
\left(
\begin{array}{c}
 x\\
 y\\
\end{array}
\right)
+ST
=
\left(
\begin{array}{cc}
\alpha_{1}& \alpha_{2}\\
\beta_{1}& \beta_{2}\\
\end{array}
\right)\cdot
\left(
\begin{array}{c}
 x\\
 y\\
\end{array}
\right)
+
\left(
\begin{array}{c}
 O_{x}\\
 O_{y}\\
\end{array}
\right)
\end{equation}
\begin{equation}
\left(
\begin{array}{c}
R^{'}(x,y)\\
G^{'}(x,y)\\
B^{'}(x,y)\\
\end{array}
\right)
=CA \cdot
\left(
\begin{array}{c}
R(x,y)\\
G(x,y)\\
B(x,y)\\
\end{array}
\right)
+CT
=
\left(
\begin{array}{ccc}
\ a_{1}&a_{2}&a_{2}\\
\ b_{1}&b_{2}&b_{2}\\
\ c_{1}&c_{2}&c_{2}\\
\end{array}
\right)\cdot
\left(
\begin{array}{c}
R(x,y)\\
G(x,y)\\
B(x,y)\\
\end{array}
\right)
+
\left(
\begin{array}{c}
O_{R}\\
O_{G}\\
O_{B}\\
\end{array}
\right)
\end{equation}
where SA and CA are nonsingular matrices.

For the color image $I(R(x,y),G(x,y),B(x,y))$, let $(x_{p},y_{p}),(x_{q},y_{q}),(x_{r},y_{r})$ be three arbitrary points in the domain of $I$. The shape primitive and the color primitive are defined by
\begin{equation}
S(p,q)=
\left|
\begin{array}{cc}
(x_{p}-\bar{x})&(x_{q}-\bar{x})\\
(y_{p}-\bar{y})&(y_{q}-\bar{y})\\
\end{array}
\right|
\end{equation}
\begin{equation}
\begin{split}
&C(p,q,r)=
\left|
\begin{array}{ccc}
(R(x_{p},y_{p})-\bar{R})&(R(x_{q},y_{q})-\bar{R})&(R(x_{r},y_{r})-\bar{R})\\
(G(x_{p},y_{p})-\bar{G})&(G(x_{q},y_{q})-\bar{G})&(G(x_{r},y_{r})-\bar{G})\\
(B(x_{p},y_{p})-\bar{B})&(B(x_{q},y_{q})-\bar{B})&(B(x_{r},y_{r})-\bar{B})\\
\end{array}
\right|
\end{split}
\end{equation}
where $\bar{A}$ represents the mean value of $A$, $A \in \{x,y,R,G,B\}$.

Then, using Eq.(3) and (4), the shape core can be defined by
\begin{equation}
sCore(n,m;d_{1},d_{2},...,d_{n})=\underbrace{S(1,2)S(k,l)...S(r,n)}_m\
\end{equation}
where n and m represent that the sCore is the product of m shape primitives which are constructed by N points $(x_{1},y_{1}),(x_{2},y_{2}),...,(x_{n},y_{n})$. $k<l$, $r<n$, $k,l,r \in \left\{1,2,...n\right\}$. $d_{i}$ represents the number of point $(x_{i},y_{i})$ in all shape primitives, $i=1,2,...,n$.

Similarly, the color core can be defined by
\begin{equation}
cCore(N,M;D_{1},D_{2},...,D_{N})=\underbrace{C(1,2,3)C(G,K,L)...C(P,Q,N)}_M\
\end{equation}
Where N and M represent that the cCore is the product of M color primitives which are constructed by N points $(x_{1},y_{1}),(x_{2},y_{2}),...,(x_{N},y_{N})$. $G<K<L$, $P<Q<N$, $G,K,L,P,Q \in \left\{1,2,...N\right\}$. $D_{i}$ represents the number of point $(x_{i},y_{i})$ in all color primitives, $i=1,2,...,N$.

Suppose the color image $I(R(x,y),G(x,y),B(x,y))$ is transformed into the image $I^{'}(R^{'}(x^{'},y^{'}),G^{'}(x^{'},y^{'}),B^{'}(x^{'},y^{'}))$ by two transformations defined by Eq.(1) and (2), $(x^{'}_{p},y^{'}_{p}),(x^{'}_{q},y^{'}_{q}),(x^{'}_{r},y^{'}_{r})$ in $I^{'}$ are the corresponding points of $(x_{p},y_{p}),(x_{q},y_{q}),(x_{r},y_{r})$ in $I$. Gong et al.\cite{14} have proved
\begin{equation}
 S^{'}(p,q)=|SA| \cdot S(p,q)
\end{equation}
\begin{equation}
 C^{'}(p,q,r)=|CA| \cdot C(p,q,r)
\end{equation}

Further results can be concluded
\begin{equation}
  sCore^{'}(n,m;d_{1},d_{2},...,d_{n})=|SA|^{m}\cdot sCore(n,m;d_{1},d_{2},...,d_{n})
\end{equation}
\begin{equation}
  cCore^{'}(N,M;D_{1},D_{2},...,D_{N})=|CA|^{M}\cdot cCore(N,M;D_{1},D_{2},...,D_{N})
\end{equation}

Therefore, the SCAMIs are constructed by
\begin{equation}
\begin{split}
&SCAMIs(n,m,N,M;d_{1},...,d_{n};D_{1},...,D_{N})\\&=
\frac {I(sCore(n,m,d_{1},...,d_{n})\cdot cCore(N,M,D_{1},...,D_{N}))}{I(sCore(1,0))^{max(n+N)+m-\frac{3M}{2}}\cdot I(cCore(3,2;2,2,2))^{\frac{M}{2}}}\\
\end{split}
\end{equation}

Then there is a relation
\begin{equation}
\begin{split}
 SCAMIs^{'}(n,m,N,M;d_{1},...,d_{n};D_{1},...,D_{N})&\\=SCAMIs(n,m,N,M;d_{1},...,d_{n};D_{1},...,D_{N})
\end{split}
\end{equation}

It must be said that $\max\{n,N\}$, $\max\limits_{i}\{d_{i}\}$ and $\max\limits_{i}\{D_{i}\}$ are named the degree, the shape order and the color order of SCAMIs, respectively. In fact, Eq.(11) can be expressed as polynomial of shape-color moment. This moment was first proposed in \cite{16} and defined by
\begin{equation}
SCM_{pq\alpha \beta \gamma}=\iint (x-\bar{x})^{p}(y-\bar{y})^{q}(R(x,y)-\bar{R})^{\alpha}(G(x,y)-\bar{G})^{\beta}(B(x,y)-\bar{B})^{\gamma}dxdy
\end{equation}

 Gong et al.\cite{14} proposed that they constructed all SCAMIs of which degrees $\leqslant 4$, shape orders $\leqslant 4$ and color orders $\leqslant 2$. They obtained 24 SCAMIs which are functional independencies using the method proposed by Brown \cite{17}. However, we will point out in the Section 3 that they omitted many simple and well-behaved SCAMIs.

\section{The Construction Framework of SCDMIs}
In this section, we introduce the general definitions of shape-color differential moment and shape-color primitive, firstly. Then, using the shape-color primitive, the shape-color core can be constructed. Finally, according to Eq.(11) and the shape-color core, we obtain the general construction formula of SCDMIs. Also, 50 instances of SCDMIs are given for experiments in the Section 4.
\subsection{The Definition of The General Shape-Color Moment}
\textbf{Definition 1.}

 Suppose the color image $I(R(x,y), G(x,y), B(x,y))$ have the k-order partial derivatives $(k=0,1,2,...)$. The general shape-color differential moment is defined by
 \begin{equation}
 \begin{split}
 SCM^{k}_{pq\alpha \beta \gamma}=\iint &(x-\bar{x})^{p}(y-\bar{y})^{q}(R^{(k)}(x,y)-\bar{R}^{\delta(k)})^{\alpha}(G^{(k)}(x,y)-\bar{G}^{\delta(k)})^{\beta}\\&(B^{(k)}(x,y)-\bar{B}^{\delta(k)})^{\gamma}dxdy\\
 \end{split}
 \end{equation}
where $(R^{(k)}(x,y)$, $G^{(k)}(x,y), B^{(k)}(x,y))$ represent the k-order partial derivatives of $(R(x,y), G(x,y), B(x,y))$. $\bar{R}, \bar{G}, \bar{B}$ represent the mean values of $R, G, B$. $\delta(k)$ is the impact function which is defined by
\begin{equation}
\delta(k)=\left\{
\begin{aligned}
1 &~& (k=0) \\
0 &~&  (k\neq 0)\\
\end{aligned}
\right.
\end{equation}

We can find that Eq.(14) and Eq.(13) are identical, when $k=0$. Therefore, the shape-color moment is a special case of the general shape-color differential moment.

\subsection{The Construction of The General Shape-Color Primitive}
\textbf{Definition 2.}

Suppose the color image $I(R(x,y), G(x,y), B(x,y))$ have the k-order partial derivatives $(k=0,1,2,...)$. $(x_{p},y_{p}),(x_{q},y_{q}),(x_{r},y_{r})$ are three arbitrary points in the domain of $I$. The general shape-color primitive is defined by
\begin{equation}
\begin{split}
SCP_{k}(p,q,r)=
\left|
\begin{array}{ccc}
F^{k}_{R}(x_{p},y_{p})&F^{k}_{R}(x_{q},y_{q})&F^{k}_{R}(x_{r},y_{r})\\
F^{k}_{G}(x_{p},y_{p})&F^{k}_{G}(x_{q},y_{q})&F^{k}_{G}(x_{r},y_{r})\\
F^{k}_{B}(x_{p},y_{p})&F^{k}_{B}(x_{q},y_{q})&F^{k}_{B}(x_{r},y_{r})\\
\end{array}
\right|
\end{split}
\end{equation}
where
\begin{equation}
 F^{k}_{C}(x,y)=\sum_{i=0}^{k}\binom{k}{i}(x-\bar{x})^{i}(y-\bar{y})^{k-i}\frac{\partial^{k}C(x,y)}{\partial x^{i}\partial y^{k-i}},
 ~~~C \in \{R,G,B\}.
\end{equation}

We can find that $C(p,q,r)$ defined by Eq.(4) is a special case of $SCP_{k}(p,q,r)$, when $k=0$.
\subsection{The Construction of The General Shape-Color Core}
\textbf{Definition 3.}

Using Definition 2, the general shape-color core is defined by
\begin{equation}
scCore_{k}(N,M;D_{1},D_{2},...,D_{N})=\underbrace{SCP_{k}(1,2,3)SCP_{k}(G,K,L)...SCP_{k}(P,Q,N)}_M\
\end{equation}
Where $k=1,2,...$, N and M represent that the $scCore_{k}$ is the product of M shape-color primitives constructed by N points $(x_{1},y_{1}),(x_{2},y_{2}),...,(x_{N},y_{N})$. $G<K<L$, $P<Q<N$, $G,K,L,P,Q \in \left\{1,2,...N\right\}$. $D_{i}$ represents the number of point $(x_{i},y_{i})$ in all shape-color primitives, $i=1,2,...,N$.

Obviously, $cCore(N,M;D_{1},D_{2},...,D_{N})$ defined by Eq.(6) is a special case of $scCore_{k}(N,M;D_{1},D_{2},...,D_{N})$, when $k=0$.
\subsection{The Construction of SCDMIs}
\textbf{Theorem 1.}

Let the color image $I(R(x,y),G(x,y),B(x,y))$ be transformed into the image $I^{'}(R^{'}(x^{'},y^{'}),G^{'}(x^{'},y^{'}),B^{'}(x^{'},y^{'}))$ by Eq.(1) and Eq.(2), $(x_{p}^{'},y_{p}^{'}),(x_{q}^{'},y_{q}^{'}),(x_{r}^{'},y_{r}^{'})$ in $I^{'}$ are corresponding points of $(x_{p},y_{p}),(x_{q},y_{q}),(x_{r},y_{r})$ in $I$, respectively. Suppose that $R(x,y), G(x,y), B(x,y), R^{'}(x^{'},y^{'}), G^{'}(x^{'},y^{'}), B^{'}(x^{'},y^{'})$ have the k-order partial derivatives $(k=0,1,2,...)$. Then there is a relation
\begin{equation}
SCP_{k}^{'}(p,q,r)=|CA| \cdot SCP_{k}(p,q,r)
\end{equation}
where
\begin{equation}
\begin{split}
SCP_{k}^{'}(p,q,r)=
\left|
\begin{array}{ccc}
F^{k}_{R^{'}}(x_{p}^{'},y_{p}^{'})&F^{k}_{R^{'}}(x_{q}^{'},y_{q}^{'})&F^{k}_{R^{'}}(x_{r}^{'},y_{r}^{'})\\
F^{k}_{G^{'}}(x_{p}^{'},y_{p}^{'})&F^{k}_{G^{'}}(x_{q}^{'},y_{q}^{'})&F^{k}_{G^{'}}(x_{r}^{'},y_{r}^{'})\\
F^{k}_{B^{'}}(x_{p}^{'},y_{p}^{'})&F^{k}_{B^{'}}(x_{q}^{'},y_{q}^{'})&F^{k}_{B^{'}}(x_{r}^{'},y_{r}^{'})\\
\end{array}
\right|
\end{split}
\end{equation}
Further, the following relation can be obtained
\begin{equation}
  scCore_{k}^{'}(N,M;D_{1},D_{2},...,D_{N})=|CA|^{M} \cdot scCore_{k}(N,M;D_{1},D_{2},...,D_{N})
\end{equation}
where
\begin{equation}
  scCore_{k}^{'}(N,M;D_{1},D_{2},...,D_{N})=\underbrace{SCP_{k}^{'}(1,2,3)SCP_{k}^{'}(G,K,L)...SCP_{k}^{'}(P,Q,N)}_M\
\end{equation}

By using Maple2015, the proof of Theorem 1 is obvious. We can find that Eq.(10) is a special case of Eq.(20), when $k=0$. So, when we replace $cCore(N,M;\\D_{1},D_{2},...,D_{N})$ in Eq.(11) with $scCore_{k}(N,M;D_{1},D_{2},...,D_{N})$, Eq.(12) is still tenable. Now, we can define SCDMIs.

\hspace{-0.25in}\textbf{Theorem 2.}
\begin{equation}
\begin{split}
&SCDMIs_{k}(n,m,N,M;d_{1},...,d_{n};D_{1},...,D_{N})\\&=
\frac {I(sCore(n,m,d_{1},...,d_{n})\cdot scCore_{k}(N,M,D_{1},...,D_{N}))}{I(sCore(1,0))^{max(n+N)+m-\frac{3M}{2}}\cdot I(scCore_{k}(3,2;2,2,2))^{\frac{M}{2}}}\\
\end{split}
\end{equation}
Then there is a relation
\begin{equation}
\begin{split}
  SCDMIs_{k}^{'}(n,m,N,M;d_{1},...,d_{n};D_{1},...,D_{N})&\\=SCDMIs_{k}(n,m,N,M;d_{1},...,d_{n};D_{1},...,D_{N})\
\end{split}
\end{equation}
where
\begin{equation}
\begin{split}
&SCDMIs_{k}^{'}(n,m,N,M;d_{1},...,d_{n};D_{1},...,D_{N})\\&=
\frac {I(sCore^{'}(n,m,d_{1},...,d_{n})\cdot scCore_{k}^{'}(N,M,D_{1},...,D_{N}))}{I(sCore^{'}(1,0))^{max(n+N)+m-\frac{3M}{2}}\cdot I(scCore_{k}^{'}(3,2;2,2,2))^{\frac{M}{2}}}\\
\end{split}
\end{equation}

Eq.(23) can be expressed as polynomial of $SCM^{k}_{pq\alpha \beta \gamma}$. Eq.(11) is a special case of Eq.(23) when $k=0$. The proof of Eq.(24) is exactly the same as that of Eq.(12) proposed in \cite{14}.

\subsection{The Instances of SCDMIs}

We can use Eq.(24) to construct instances of $SCDMIs$ by setting different k values. However, color images are discrete data, the partial derivatives of each order can't be accurately calculated. With the elevation of order, the error will be more and more, which will greatly affect the stability of SCDMIs. So, we set $k=0,1$ in this paper.

When $k=0$, $SCDMIs_{0}$ are equivalent to SCAMIs. We construct $SCDMIs_{0}$ of which degrees $\leqslant 4$, shape orders $\leqslant 4$ and color orders $\leqslant 1$. Gong et al. \cite{14} thought that in order to obtain the $SCDMIs_{0}$ of which degrees $\leqslant 4$, shape orders $\leqslant 4$ and color orders $\leqslant 1$, $scCore_{0}(3,1;1,1,1)$ must be $C(1,2,3)$. This judgment is wrong. In fact, $scCore_{0}(3,1;1,1,1)$ can be $C(1,2,3)$, $C(1,2,4)$, $C(1,3,4)$ and $C(2,3,4)$. Thus, lots of $SCDMIs_{0}$ were missed in \cite{14}. By correcting this shortcoming, we get 25 $SCDMIs_{0}$ that satisfy the independence of the function by using the method proposed by \cite{17}.

At the same time, when $k=1$, $SCP_{1}(p,q,r)$ is defined by
\begin{equation}
 SCP_{1}(p,q,r)=
 \left|
\begin{array}{ccc}
F^{1}_{R}(x_{p},y_{p})&F^{1}_{R}(x_{q},y_{q})&F^{1}_{R}(x_{r},y_{r})\\
F^{1}_{G}(x_{p},y_{p})&F^{1}_{G}(x_{q},y_{q})&F^{1}_{G}(x_{r},y_{r})\\
F^{1}_{B}(x_{p},y_{p})&F^{1}_{B}(x_{q},y_{q})&F^{1}_{B}(x_{r},y_{r})\\
\end{array}
\right|
\end{equation}

where
\begin{equation}
 F^{1}_{C}(x,y)=(x-\bar{x})\frac{\partial C}{\partial x}+(y-\bar{y})\frac{\partial C}{\partial y}
 ~~~C \in \{R,G,B\}.
\end{equation}

By replacing $C(1,2,3)$, $C(1,2,4)$, $C(1,3,4)$ and $C(2,3,4)$ with $SCP_{1}(1,2,3)$, $SCP_{1}(1,2,4)$, $SCP_{1}(1,3,4)$ and $SCP_{1}(2,3,4)$, 25 $SCDMIs_{1}$ can be obtained. Therefore, we can construct the feature vector SCDMI50, which is defined by
\begin{equation}
  SCDMI50=[SCDMIs_{0}^{1},...,SCDMIs_{0}^{25},SCDMIs_{1}^{1},...,SCDMIs_{1}^{25}]
\end{equation}

The construction methods of 50 instances are shown in Table 1.
\begin{table}[!h]
\caption{The construction methods of $SCDMI50$ }
\newcommand{\tabincell}[2]{\begin{tabular}{@{}#1@{}}#2\end{tabular}}
\begin{tabular}{lll}
  \hline
  Name & ~~scCore  & ~~ sCore \\
  \hline
  $SCDMIs_{0}^{1}/SCDMIs_{1}^{1}$ & ~~$C(1,2,3)/SCP_{1}(1,2,3)$ &~~$(x_{1}y_{2}-x_{2}y_{1})(x_{1}y_{3}-x_{3}y_{1})^{2}$ \\
  $SCDMIs_{0}^{2}/SCDMIs_{1}^{2}$ & ~~$C(1,2,3)/SCP_{1}(1,2,3)$ &~~$(x_{1}y_{2}-x_{2}y_{1})(x_{1}y_{3}-x_{3}y_{1})^{3}$ \\
  $SCDMIs_{0}^{3}/SCDMIs_{1}^{3}$ & ~~$C(1,2,3)/SCP_{1}(1,2,3)$ &~~$(x_{1}y_{2}-x_{2}y_{1})(x_{1}y_{3}-x_{3}y_{1})(x_{2}y_{3}-x_{3}y_{2})$ \\
  $SCDMIs_{0}^{4}/SCDMIs_{1}^{4}$ & ~~$C(1,2,3)/SCP_{1}(1,2,3)$ &~~$(x_{1}y_{2}-x_{2}y_{1})(x_{1}y_{3}-x_{3}y_{1})(x_{2}y_{3}-x_{3}y_{2})^{3}$\\
  $SCDMIs_{0}^{5}/SCDMIs_{1}^{5}$ & ~~$C(1,2,3)/SCP_{1}(1,2,3)$ &~~$(x_{1}y_{2}-x_{2}y_{1})^{2}(x_{1}y_{3}-x_{3}y_{1})^{2}(x_{2}y_{3}-x_{3}y_{2})$ \\
  $SCDMIs_{0}^{6}/SCDMIs_{1}^{6}$ & ~~$C(1,2,4)/SCP_{1}(1,2,4)$ &~~$(x_{1}y_{2}-x_{2}y_{1})(x_{2}y_{3}-x_{3}y_{2})(x_{3}y_{4}-x_{4}y_{3})$ \\
  $SCDMIs_{0}^{7}/SCDMIs_{1}^{7}$ & ~~$C(1,2,4)/SCP_{1}(1,2,4)$ &~~$(x_{1}y_{2}-x_{2}y_{1})(x_{2}y_{3}-x_{3}y_{2})(x_{3}y_{4}-x_{4}y_{3})^{3}$\\
  $SCDMIs_{0}^{8}/SCDMIs_{1}^{8}$ & ~~$C(1,3,4)/SCP_{1}(1,3,4)$ &~~$(x_{1}y_{2}-x_{2}y_{1})(x_{2}y_{3}-x_{3}y_{2})(x_{3}y_{4}-x_{4}y_{3})^{3}$\\
  $SCDMIs_{0}^{9}/SCDMIs_{1}^{9}$ & ~~$C(1,2,3)/SCP_{1}(1,2,3)$ &~~$(x_{1}y_{2}-x_{2}y_{1})^{2}(x_{2}y_{3}-x_{3}y_{2})(x_{3}y_{4}-x_{4}y_{2})$\\
  $SCDMIs_{0}^{10}/SCDMIs_{1}^{10}$ &~~$C(1,2,4)/SCP_{1}(1,2,4)$ &~~$(x_{1}y_{2}-x_{2}y_{1})^{2}(x_{2}y_{3}-x_{3}y_{2})(x_{3}y_{4}-x_{4}y_{3})^{3}$ \\
  $SCDMIs_{0}^{11}/SCDMIs_{1}^{11}$ &~~$C(2,3,4)/SCP_{1}(2,3,4)$ &~~$(x_{1}y_{2}-x_{2}y_{1})^{2}(x_{2}y_{3}-x_{3}y_{2})(x_{3}y_{4}-x_{4}y_{3})^{3}$\\
  $SCDMIs_{0}^{12}/SCDMIs_{1}^{12}$ &~~$C(1,2,4)/SCP_{1}(1,2,4)$ &~~$(x_{1}y_{2}-x_{2}y_{1})^{3}(x_{2}y_{3}-x_{3}y_{2})(x_{3}y_{4}-x_{4}y_{3})^{3}$\\
  $SCDMIs_{0}^{13}/SCDMIs_{1}^{13}$ &~~$C(1,2,3)/SCP_{1}(1,2,3)$ &~~$(x_{1}y_{2}-x_{2}y_{1})(x_{2}y_{3}-x_{3}y_{2})^{2}(x_{3}y_{4}-x_{4}y_{3})^{2}$\\
  $SCDMIs_{0}^{14}/SCDMIs_{1}^{14}$ &~~$C(1,2,4)/SCP_{1}(1,2,4)$ &~~$(x_{1}y_{2}-x_{2}y_{1})(x_{2}y_{3}-x_{3}y_{2})^{2}(x_{3}y_{4}-x_{4}y_{3})^{2}$ \\
  $SCDMIs_{0}^{15}/SCDMIs_{1}^{15}$ &~~$C(1,2,4)/SCP_{1}(1,2,4)$ &~~$(x_{1}y_{2}-x_{2}y_{1})(x_{2}y_{3}-x_{3}y_{2})^{3}(x_{3}y_{4}-x_{4}y_{3})$\\
  $SCDMIs_{0}^{16}/SCDMIs_{1}^{16}$ &~~$C(1,3,4)/SCP_{1}(1,3,4)$ &~~\tabincell{l}{$(x_{1}y_{2}-x_{2}y_{1})(x_{2}y_{3}-x_{3}y_{2})(x_{3}y_{4}-x_{4}y_{3})^{2}\cdot$\\$(x_{4}y_{1}-x_{1}y_{4})$}\\
  $SCDMIs_{0}^{17}/SCDMIs_{1}^{17}$ &~~$C(1,2,3)/SCP_{1}(1,2,3)$ &~~\tabincell{l}{$(x_{1}y_{2}-x_{2}y_{1})^{2}(x_{2}y_{3}-x_{3}y_{2})(x_{3}y_{4}-x_{4}y_{3})^{3}\cdot$\\$(x_{4}y_{1}-x_{1}y_{4})$}\\
  $SCDMIs_{0}^{18}/SCDMIs_{1}^{18}$ &~~$C(1,2,4)/SCP_{1}(1,2,4)$ &~~$(x_{1}y_{2}-x_{2}y_{1})(x_{1}y_{3}-x_{3}y_{1})(x_{1}y_{4}-x_{4}y_{1})$\\
  $SCDMIs_{0}^{19}/SCDMIs_{1}^{19}$ &~~$C(1,2,4)/SCP_{1}(1,2,4)$ &~~\tabincell{l}{$(x_{1}y_{2}-x_{2}y_{1})(x_{1}y_{3}-x_{3}y_{1})(x_{1}y_{4}-x_{4}y_{1})\cdot$\\$(x_{3}y_{4}-x_{4}y_{3})^{3}$} \\
  $SCDMIs_{0}^{20}/SCDMIs_{1}^{20}$ &~~$C(2,3,4)/SCP_{1}(2,3,4)$ &~~\tabincell{l}{$(x_{1}y_{2}-x_{2}y_{1})(x_{1}y_{3}-x_{3}y_{1})^{2}(x_{1}y_{4}-x_{4}y_{1})\cdot$\\$(x_{3}y_{4}-x_{4}y_{3})^{2}$}\\
  $SCDMIs_{0}^{21}/SCDMIs_{1}^{21}$ &~~$C(1,2,3)/SCP_{1}(1,2,3)$ &~~\tabincell{l}{$(x_{1}y_{2}-x_{2}y_{1})^{2}(x_{1}y_{3}-x_{3}y_{1})(x_{1}y_{4}-x_{4}y_{1})\cdot$\\$(x_{3}y_{4}-x_{4}y_{3})$}\\
  $SCDMIs_{0}^{22}/SCDMIs_{1}^{22}$ &~~$C(1,2,3)/SCP_{1}(1,2,3)$ &~~\tabincell{l}{$(x_{1}y_{2}-x_{2}y_{1})^{2}(x_{1}y_{3}-x_{3}y_{1})(x_{1}y_{4}-x_{4}y_{1})\cdot$\\$(x_{3}y_{4}-x_{4}y_{3})^{3}$}\\
  $SCDMIs_{0}^{23}/SCDMIs_{1}^{23}$ &~~$C(1,2,4)/SCP_{1}(1,2,4)$ &~~\tabincell{l}{$(x_{1}y_{2}-x_{2}y_{1})^{2}(x_{1}y_{3}-x_{3}y_{1})(x_{1}y_{4}-x_{4}y_{1})\cdot$\\$(x_{3}y_{4}-x_{4}y_{3})^{3}$}\\
  $SCDMIs_{0}^{24}/SCDMIs_{1}^{24}$ &~~$C(1,2,4)/SCP_{1}(1,2,4)$ &~~\tabincell{l}{$(x_{1}y_{2}-x_{2}y_{1})(x_{2}y_{3}-x_{3}y_{2})(x_{3}y_{4}-x_{4}y_{3})^{2}\cdot$\\$(x_{4}y_{1}-x_{1}y_{4})(x_{2}y_{4}-x_{4}y_{2})$}\\
  $SCDMIs_{0}^{25}/SCDMIs_{1}^{25}$ &~~$C(1,2,4)/SCP_{1}(1,2,4)$ &~~\tabincell{l}{$(x_{1}y_{2}-x_{2}y_{1})^{2}(x_{2}y_{3}-x_{3}y_{2})(x_{3}y_{4}-x_{4}y_{3})^{2}\cdot$\\$(x_{4}y_{1}-x_{1}y_{4})^{2}(x_{2}y_{4}-x_{4}y_{2})$}\\
  \hline
\end{tabular}
\end{table}
In order to more clearly explain that SCDMIs can be expanded into the polynomial of $SCM^{k}_{pq\alpha \beta \gamma}$, we give the shape-color moment polynomial of $SCMIs_{0}^{3}$ which is represented by
\begin{equation}
\begin{split}
  SCDMI^{3}_{0}=&\{6SCM^{0}_{02001}SCM^{0}_{11010}SCM^{0}_{20100}-6SCM^{0}_{02001}SCM^{0}_{11100}SCM^{0}_{20010}\\&-6SCM^{0}_{02010}SCM^{0}_{11001}SCM^{0}_{20100}+6SCM^{0}_{02010}SCM^{0}_{11100}SCM^{0}_{20001}\\&+6SCM^{0}_{02100}SCM^{0}_{11001}SCM^{0}_{20010}-6SCM^{0}_{02100}SCM^{0}_{11010}SCM^{0}_{20001}\}\\&/\{6SCM^{0}_{00002} SCM^{0}_{00020} SCM^{0}_{00200}-6 SCM^{0}_{00002} (SCM^{0}_{00110})^{2}\\&-6 (SCM^{0}_{00011})^{2} SCM^{0}_{00200}+12 SCM^{0}_{00011} SCM^{0}_{00101} SCM^{0}_{00110}\\&-6 SCM^{0}_{00020} (SCM^{0}_{00101})^{2}\}\\
\end{split}
\end{equation}

\section{Experimental Results}
In this section, some experiments are designed to evaluate the performance of SCDMI50. Firstly, we verify the stability and discriminability  of SCDMI50 by using synthetic images. Then, some retrieval experiments based on real image databases are performed. Also, we chose some commonly used image descriptors for comparison.

It is worth noting that we have to choose the method to calculate the first order partial differentials of $R(x,y)$, $G(x,y)$ and $B(x,y)$. The 5 points difference formulas are used for approximating the first partial derivatives of discrete color images, which are defined by
\begin{equation}
\begin{split}
  &\frac {\partial C(x,y)}{\partial x}=C(x-2,y)-8C(x-1,y)+8C(x+1,y)-C(x+2,y)\\
  &\frac {\partial C(x,y)}{\partial y}=C(x,y-2)-8C(x,y-1)+8C(x,y+1)-C(x,y+2)\\
\end{split}
\end{equation}
where $C \in \{R,G,B\}$. We choose this method because it guarantees the computational accuracy of the first partial differential to a certain extent and also maintains a relatively fast calculation speed.
\subsection{The Stability and Discriminability of SCDMI50}
We select 50 different kinds of butterfly images, which are shown in the Fig.(1.a). Then, 5 shape affine transforms and 4 color affine transforms are applied to each image. So, one image can get 20 transformed versions which are shown in the Fig.(1.b). Thus, we obtain the database containing 1000 images. $10\%$ images are used as the training data and the rest make up the testing data.
\begin{figure}[H]
  \centering
  \subfigure[]{
  \label{Fig.sub.1}
  \includegraphics[width=100mm,height=38mm]{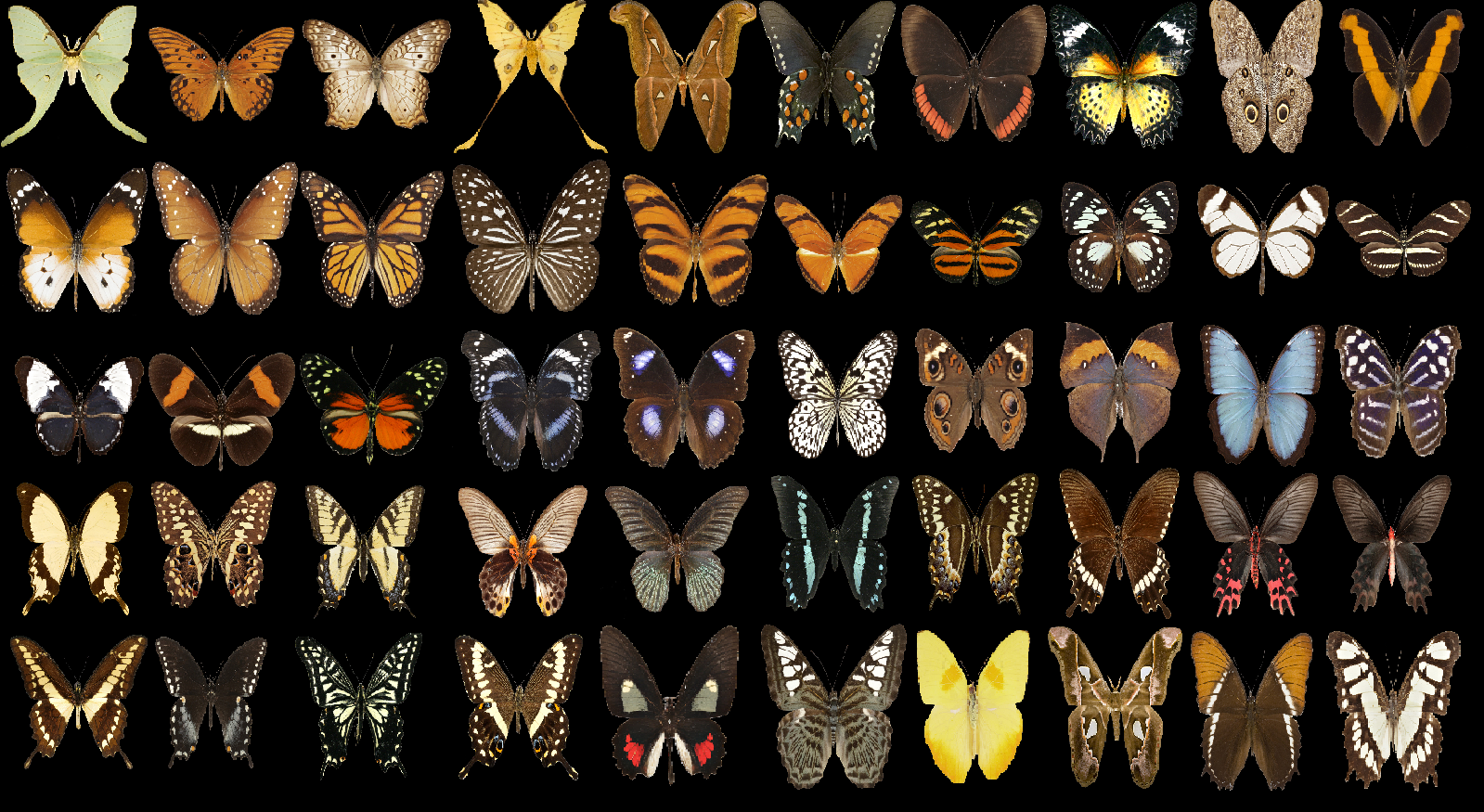}}
    \subfigure[]{
  \label{Fig.sub.2}
  \includegraphics[width=100mm,height=38mm]{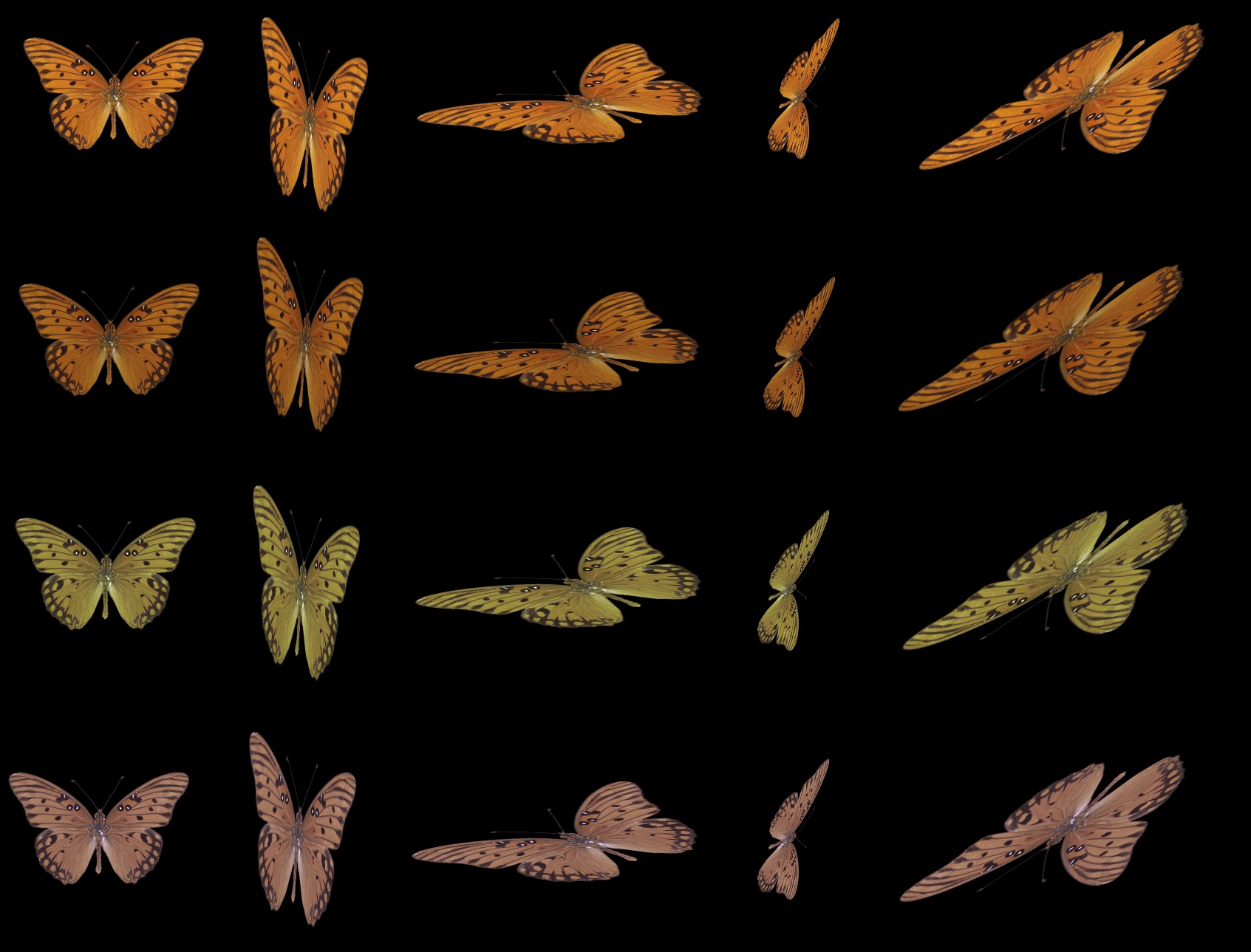}}
  \caption{(a) 50 different kinds of butterfly images; (b) 20 transformed versions of the image.}
  \label{1}
\end{figure}
For comparison with $SCDMI50$, we chose some commonly used features of gray images or color images.
\begin{itemize}
  \item \textbf{Hu moments}, which were composed of 7 invariants under the shape similarity transform, proposed in \cite{1}. Hu moments were designed for gray images.
  \item \textbf{AMIs} which were composed of 17 invariants under the shape affine transform proposed in \cite{18}. AMIs were designed for gray images.
  \item \textbf{RGhistogram}, which consisted of 60-dimensional features and was scale-invariant in color space, proposed in \cite{19}.
  \item \textbf{Transformed color distribution}, which consisted of 60-dimensional features, was proposed in \cite{19}. Transformed color distribution was scale-invariant and sift-invariant in color space.
  \item \textbf{Color moments} consisted of the first, second and third geometric moments of each channel of the color image, which were proposed in \cite{15}. Color moments are sift-invariant in color space.
  \item \textbf{GPSOs} were invariant under the shape affine transforms and the color diagonal-offset transform, which were proposed in \cite{10}. GPSOs consisted of 21 moment invariants.
\end{itemize}

Subsequently, image classification is performed on the butterfly database by using different kind of features. The Chi-Square distance is selected to measure the similarity of two images. Finally, we plot the classification accuracies obtained by using different features, which are shown in the Table 2.

\begin{table}[!h]
\centering
\caption{The classification accuracies obtained by using different features}
\begin{tabular}{c|c|c|c}
  \hline
  Descriptor & ~~Accuracy &~~Descriptor & ~~Accuracy \\
  \hline
  SCDMI50 & ~~\textbf{98.67\%} &~~GPSOs & ~~78.56\% \\
  \hline
  AMIs & ~~50.11\% &~~Hu moments & ~~25.00\% \\
  \hline
  Color moments & ~~74.77\% &~~RGhistogram & ~~80.56\% \\
  \hline
  Transformed color distribution & ~~95.78\% &&  \\
  \hline
\end{tabular}
\end{table}

On the one hand, we can find that the result obtained by using $SCDMI50$ is better than those obtained by using other features. And, color information is very important for the classification. On the other hand, the Table 2 show that $SCDMI50$ have good stability and distinguishability for the shape affine and color affine transforms, and demonstrate that the construction formula of SCDMIs designed in the Section 3 is correct.
\subsection{Image Retrieval on Real Image Database}
In order to further illustrate the performance of $SCDMI50$ better than the commonly used features listed in the Section 4.1, the database COIL-100 proposed in \cite{20} is chosen for our experiment. COIL-100 contains 7202 images of 100 categories, each of which has 72 images taken from different angles. We choose 30 classes, each class contains 5 images. For each image, 6 color affine transformations are applied. Thus, 900 images of 30 categories are obtained, each of categories contains 30 images which are shown in the Fig.(2).
\begin{figure}[H]
  \centering
  \includegraphics[width=100mm,height=50mm]{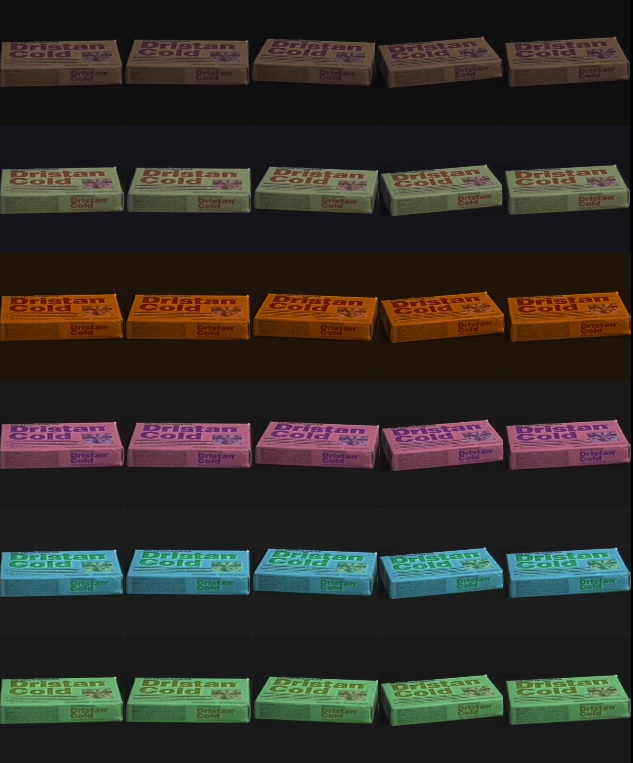}\\
  \caption{Each of categories contains 30 images}\label{1}
\end{figure}
Similar to the Section 4.1, the image retrieval experiment is performed on this database. The Precision-Recall curves are shown in the Fig.(3).

\begin{figure}
  \centering
  \includegraphics[width=100mm,height=60mm]{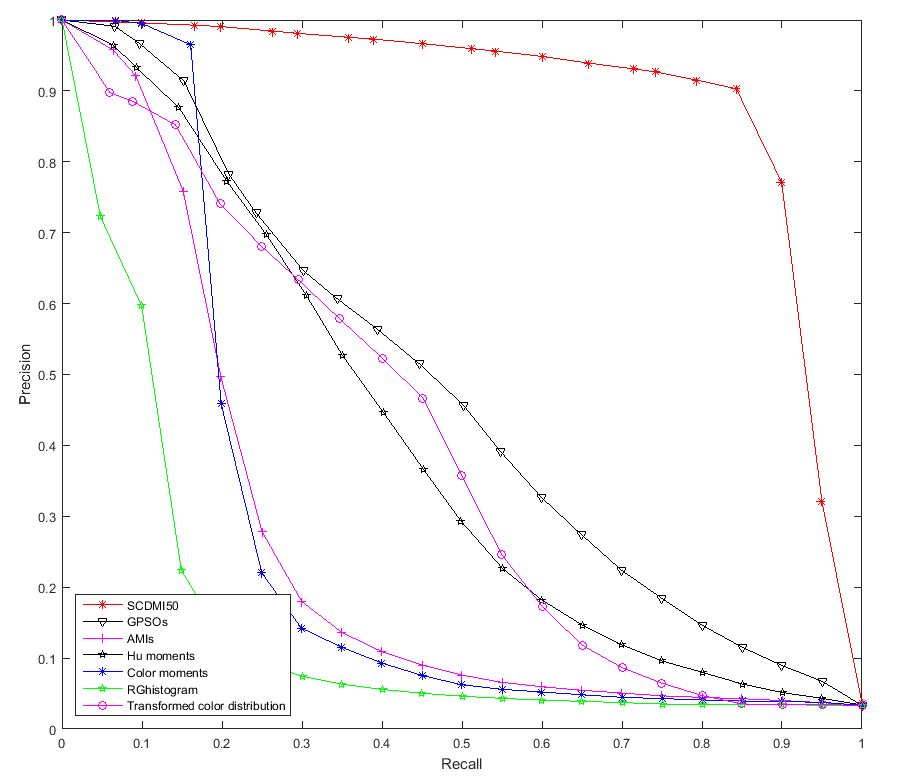}\\
  \caption{Precision-Recall curves obtained by using different features on the COIL-100 database}\label{1}
\end{figure}

Obviously, the result obtained by using SCDMI50 is far superior to those obtained by using other features. This is because that traditional image features are constructed by using only color information or shape information. However, we use two kinds of information while constructing SCDMI50. In addition, traditional image descriptors are only stable for simple changes in color space, such as the diagonal-offset transforms. When the color space changes drastically, they are less robust. 

Finally, in order to observe the performance of $SCDMIs_{0}$ and $SCDMIs_{1}$ clearly, we compare the retrieval results of $SCDMI50$, $SCDMI_{0}25$ and $SCDMI_{1}25$ which are shown in the Fig.(4). Among them, $SCDMI_{0}25$ and $SCDMI_{1}25$ are defined by
\begin{equation}
   SCDMI_{0}25=[SCDMIs_{0}^{1},SCDMIs_{0}^{2},...,SCDMIs_{0}^{25}]
\end{equation}
\begin{equation}
   SCDMI_{1}25=[SCDMIs_{1}^{1},SCDMIs_{1}^{2},...,SCDMIs_{1}^{25}]
\end{equation}
\begin{figure}[H]
  \centering
  \includegraphics[width=105mm,height=60mm]{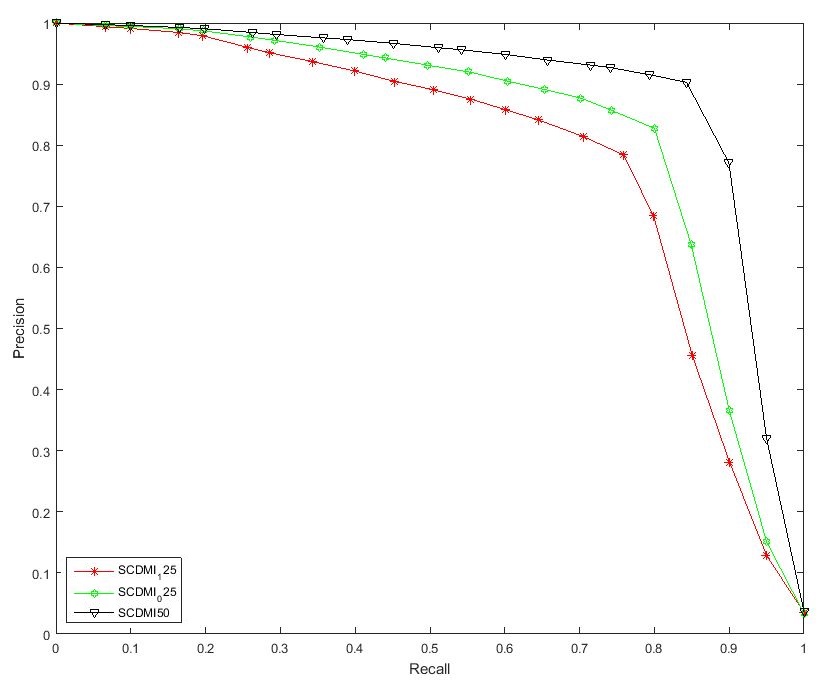}\\
  \caption{Precision-Recall curves obtained by using $SCDMI50$, $SCDMI_{0}25$ and $SCDMI_{1}25$}\label{1}
\end{figure}

Because of the error caused by the inaccuracy of partial derivative calculation, the result obtained by using $SCDMI_{1}25$ slightly worse than the result obtained $SCDMI_{0}25$, but far better than the results obtained by other commonly used features. Meanwhile, $SCDMI_{1}25$ increase the number of invariants which have simple structures and good properties. We combine $SCDMI_{1}25$ and $SCDMI_{0}25$ to get $SCDMI50$ which achieve the best retrieval result in our experiment.

\section{Conclusion}
In this paper, we proposed a kind of shape-color differential moment invariants (SCDMIs) of color images, which are invariant under the shape affine and color affine transforms, by using partial derivatives of each color channel. It is obvious that all SCAMIs proposed in \cite{14} are the special cases of SCDMIs when $k=0$. Then, we correct a mistake in \cite{14} and obtain 50 instances of SCDMIs, which have simple structures and good properties. Finally, several commonly used image descriptors and SCDMIs are used in image retrieval of color images, respectively. By comparing the experimental results, we find that SCDMIs get better results.


\begin{thebibliography}{14}
\bibitem{1} Hu, M.K.: Visual pattern recognition by moment invariants. IRE Trans. Inf. Theory 8(2), 179-187 (1962)
\bibitem{2} Zhang, Y.D., Wang, S.H., Sun, P., Phillips, P.: Pathological brain detection based on wavelet entropy and Hu moment invariants. Bio-Medical Materials and Engineering 26(s1), S1283-S1290 (2015)
\bibitem{3} Dudani, S.A., Breeding, K.J., McGhee, R.B.: Aircraft identification by moment invariants. IEEE Trans. Computers 26(1), 39-46 (1977)
\bibitem{4} Flusser, J., Suk, T.: Pattern recognition by affine moment invariants. Pattern Recognition 26(1), 167-174 (1993)
\bibitem{5} Flusser, J., Suk, T.: Affine moment invariants: a new tool for character recognition. Pattern Recognition Letters 15(4), 433-436 (1994)
\bibitem{6} Renuka, L., Vrushsen, P.: Facial Expression Recognition based on Affine Moment Invariants. International Journal of Computer Science Issues 9(6), 388-392 (2012)
\bibitem{7} Suk, T., Flusser, J.: Graph method for generating affine moment invariants. In: Proc. Internat. Conf. on Pattern Recognition, pp. 192-195. (2004)
\bibitem{8} Xu, D., Li, H.: Geometric moment invariants. Pattern Recognition 41(1), 240-249 (2008)
\bibitem{9} Geusebroek, J.M., Van den Boomgaard, R., Smeulders, A.W.M., Geerts, H.: Color Invariance. IEEE Transactions on Pattern Analysis and Machine Intelligence 23(12), 1338-1350 (2001)
\bibitem{10} Mindru, F., Tuytelaars, T., Van Gool, L., Moons, T.: Moment invariants for recognition under changing viewpoint and illumination. Comput. Vis. Image Underst. 94(1), 3-27 (2004)
\bibitem{11} Suk, T., Flusser, J.: Affine moment invariants of color images. In: International Conference on Computer Analysis of Images and Patterns, pp. 334-341 (2009)
\bibitem{12} Gong, M., Hu, P., Cao, W.G., Li, H.: A Kind of Shape-Color Moment Invariants. In: 12th International Conference on Computer-Aided Design and Computer Graphics, pp. 425-432 (2011)
\bibitem{13} Gong, M., Li, H., Cao, W.G.:  Moment invariants to affine transformation of colours. Pattern Recognition Letters 34(11), 1240-1251 (2013)
\bibitem{14} Gong, M., Hao, Y., Mo, H.L., Li, H.: Naturally Combined Shape-Color Moment Invariants under Affine Transformations, \url{http://arxiv.org/abs/1705.10928} (2017)
\bibitem{15} Stricker, M.A., Orengo, M.: Similarity of color images. IS\&T/SPIE's Symposium on Electronic Imaging: Science \& Technology, International Society for Optics and Photonics, pp. 381-392 (1995)
\bibitem{16} Mindru, F., Van Gool, L., Moons, T.: Model estimation for photometric changes of outdoor planar color surfaces caused by changes in illumination and viewpoint. In: Proc. Internat. Conf. on Pattern Recognition, pp. 620-623 (2002)
\bibitem{17} Brown, A.B.: Functional dependence. Transactions of the American Mathematical Society 38(2), 379-394 (1935)
\bibitem{18} Suk, T., Flusser, J.: Affine moment invariants generated by graph method. Pattern Recognition. 44(9), 2047-2059 (2011)
\bibitem{19} Sande, K.V.D., Gevers, T., Snoek, C.: Evaluating color Descriptors for object and scene recognition. IEEE Transactions on Pattern Analysis and Machine Intelligence 32(9), 1582-1596 (2010)
\bibitem{20} Nene, S.A., Nayer, S.K., Murase, H.: Columbia object image library($COIL-100$). Technical Report CUCS-006-96, CUCS, 1996
\end{thebibliography}
\end{document}